\documentclass[sn-mathphys]{sn-jnl}

\usepackage{epstopdf}
\jyear{2022}%

\theoremstyle{thmstyleone}%
\theoremstyle{thmstyletwo}%

\theoremstyle{thmstylethree}%

\begin{document}

\title[Parameters, Properties, and Process]{Parameters, Properties, and Process: Conditional Neural Generation of Realistic SEM Imagery Towards ML-assisted Advanced Manufacturing}


\author*[1]{\fnm{Scott} \sur{Howland}}\email{scott.howland@pnnl.gov}

\author[2]{\fnm{Lara} \sur{Kassab}}\email{lara.kassab@pnnl.gov}

\author[1]{\fnm{Keerti} \sur{Kappagantula}}\email{keertisahithi.kappagantula@pnnl.gov}

\author[2]{\fnm{Henry} \sur{Kvinge}}\email{henry.kvinge@pnnl.gov}

\author[2]{\fnm{Tegan} \sur{Emerson}}\email{tegan.emerson@pnnl.gov}

\affil*[1]{\orgname{Pacific Northwest National Laboratory}, \orgaddress{\street{902 Battelle Blvd}, \city{Richland}, \postcode{99354}, \state{WA}, \country{United States}}}

\affil*[2]{\orgname{Pacific Northwest National Laboratory}, \orgaddress{\street{1100 Dexter Ave N}, \city{Seattle}, \postcode{98109}, \state{WA}, \country{United States}}}


\abstract{The research and development cycle of advanced manufacturing processes traditionally requires a large investment of time and resources. Experiments can be expensive and are hence conducted on relatively small scales. This poses problems for typically data-hungry machine learning tools which could otherwise expedite the development cycle. We build upon prior work by applying conditional generative adversarial networks (GANs) to scanning electron microscope (SEM) imagery from an emerging advanced manufacturing process, shear assisted processing and extrusion (ShAPE). We generate realistic images \emph{conditioned} on temper and either experimental parameters or material properties. In doing so, we are able to integrate machine learning into the development cycle, by allowing a user to immediately visualize the microstructure that would arise from particular process parameters or properties. This work forms a technical backbone for a fundamentally new approach for understanding manufacturing processes in the absence of first-principle models. By characterizing microstructure from a topological perspective we are able to evaluate our models' ability to capture the breadth and diversity of experimental scanning electron microscope (SEM) samples. Our method is successful in capturing the visual and general microstructural features arising from the considered process, with analysis highlighting directions to further improve the topological realism of our synthetic imagery.}

\keywords{Deep Learning, Scanning Electron Microscopy, Conditional Image Generation, Generative Adversarial Networks}



\maketitle

\section{Introduction}
We live in an age of unprecedented technological growth. This growth has changed the daily lives of people across the globe. Changes like more compact devices, improved battery life, and faster processing are accompanied by additional methods of communication and access to/sharing of information. The former are underpinned by exploiting new materials and advanced manufacturing technologies while the latter come at the hand of increasingly sophisticated artificial intelligence (AI). Although there have been incremental advances in leveraging AI for specific analysis tasks in advanced manufacturing, there are no established, generalizable frameworks for accelerating research and development across material systems and manufacturing processes.

Because of the physical regimes in which they operate, advanced manufacturing processes are supported by nascent first-principle simulation capabilities instead of the more conventional or established approaches due to their cutting-edge nature. Owing to the ongoing emergence of such simulation or physics-based models for advanced manufacturing processes, practitioners are often forced to rely on their intuition and a small body of data when designing experiments which may or may not probe synthesis regimes that ultimately result in desires material microstructures and performance metrics. This can lead  to a slower, more expensive research and development cycle and a delay of advanced manufacturing deployment for pilot and commercial-sale applications.

It is well known in the materials science and manufacturing fields that material microstructures play a central role in associating manufacturing process parameters used in synthesizing a component (or sample) and its performance. As such, microstructural features are crucial for guiding and interpreting manufacturing data. Of the multiple methods which enable determination of metal microstructures, scanning electron microscope (SEM) imaging is a popular approach for capturing information regarding important material features such as grain size distribution, precipitate morphology, and grain boundary density amongst others. However, SEM images must be analyzed to identify key features of interest, which requires domain knowledge and post-processing activities. All this makes SEM imaging a time and resource-intensive endeavor. Accordingly, there is great interest in reducing the number of SEM images that have to be obtained for a developmental process while also decreasing the cost of associated post-processing and analysis efforts.

\begin{figure}[t]
    \centering
    \includegraphics[width=0.9\columnwidth]{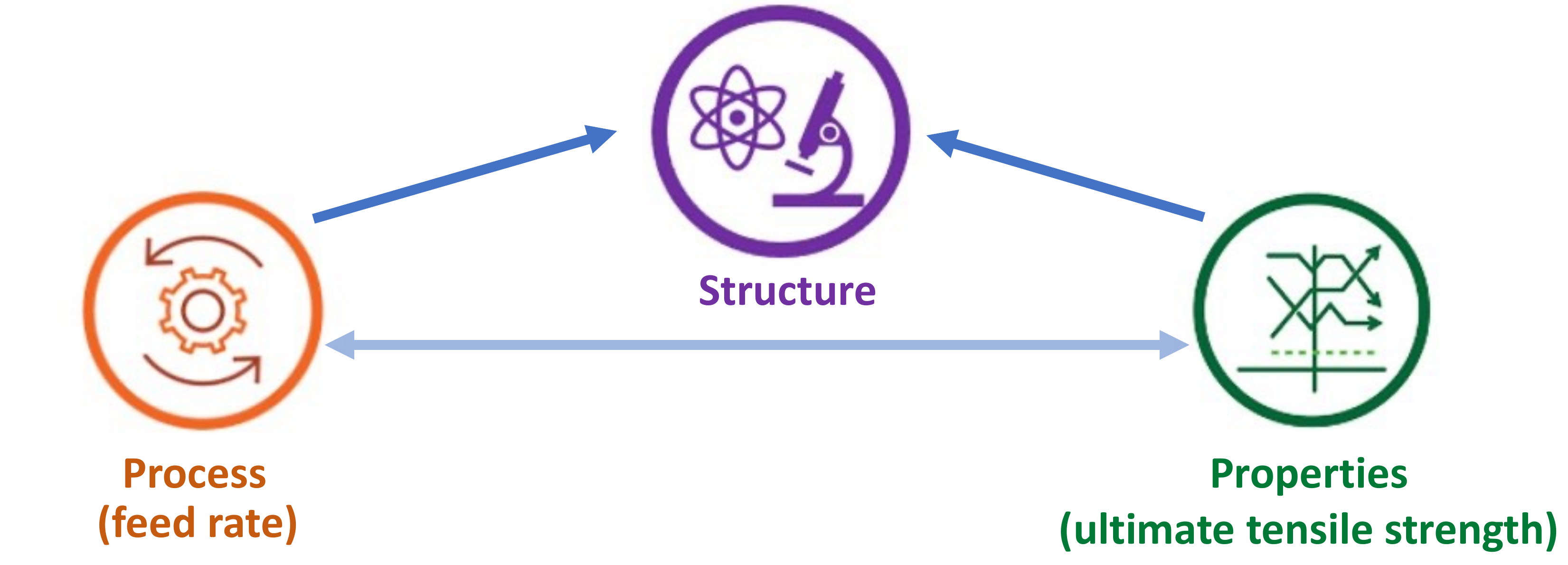}
    \caption{Process Parameter- Microstructure- Material Property Triangle. In this diagram we identify the process parameter and resulting material property we consider for ShAPE manufacturing, namely ``feed rate'' and ``ultimate tensile strength'', respectively.}
    \label{fig:triangle}
\end{figure}

There are several models available for predicting the microstructures of materials manufactured in a specific process parameter regime or identify the microstructures of the materials demonstrating a specific combination of performance metrics using first-principles approaches for conventional manufacturing approaches. However, as discussed above, such models are readily available to generate microstructures corresponding to either specific process conditions or final performance for advanced manufacturing. More recently, deep learning (DL) has found various applications to interpreting and understanding SEM images in the materials science and manufacturing applications, such as automatic classification of images \cite{azimi2018advanced, muller2020bainitic, 2020steel} and segmentation of images to identify different regions of interest \cite{durmaz2021inference}. Recently, DL methods have been used to generate SEM images of different materials \cite{iyer2019conditional} and more \cite{baskaran2021adoption}; however, it is important to note that in most of these works, the DL approach deals with images in isolation and is not explicitly informed by the manufacturing technique. It is well understood that microstructural features strongly depend on the manufacturing conditions used to produce them. Therefore, while it is a great advancement to use DL for generating SEM images, reducing the cost of associated research and development activities, we also note that these prior works are unable to incorporate the valuable, process-dependent information necessary to expedite the development-validation cycle. Subsequently, there is a critical need to generate SEM images conditioned on specific manufacturing parameters as the next wave of DL development for materials and manufacturing image analysis. Incorporating a conditional component into SEM image generation enables the production of synthetic SEM imagery which conditionally depends on either manufacturing process parameters or target material properties as illustrated in Figure~\ref{fig:triangle}. While conditional image generation models have been widely used, most DL techniques are data-hungry, which presents problems when applied to domains like materials science and advance manufacturing which suffer precisely from scarce data. Therefore, it is essential for any conditional SEM image generation to be feasible even when trained on small datasets.


In this paper, we present a rationale and approach for addressing these challenges in applying DL for SEM imagery given limited training data. We demonstrate an ability to produce realistic microstructures in synthetic images and provide methods to quantify consistency with experimental SEM images through the use of topological feature extraction. This work takes a critical step towards leveraging machine learning to help accelerate advanced manufacturing research and development in light of developing first-principles simulations. We develop generative models trained on SEM images of aluminum alloy AA7075 tubes manufactured via the Shear-Assisted Processing and Extrusion (ShAPE) technology \cite{WHALEN2021699, shaped1}. ShAPE is an emerging advanced manufacturing process that an synthesize rods, bars, tubes, and wires \cite{kalsar2022microstructureC, li2022manufactureA, li2021copperD, reza2022effectB} of different cross-sectional areas and shapes from metallic (pure metals, alloys) feedstock in various forms such as powders, chips, films, discs, and solid billets \cite{darsell2018shearG, taysom2022fabricationE, wang2020microstructuralF}. ShAPE-synthesized parts demonstrate unique microstructures with minimal porosity and never-before-seen performance. Several publications are available describing the synthesis and characterization of ShAPE samples made from aluminum, magnesium, copper, and steel, among others \cite{jiang2017frictionI, whalen2019magnesiumH}. ShAPE demonstrated enhanced performance in bulk-scale components, making their scale-up pathways relatively viable for industry. Therefore, there is an urgent need to develop models which can associate ShAPE process parameters with resulting microstructures in order to reduce research and development time and deployment delays for ShAPE at an industrial scale. 

\section{Background and Related Work}
\subsection{Manufacturing with Shear-Assisted Processing and Extrusion (ShAPE)}
This work is focused on generating SEM microstructures of aluminum alloy AA7075 tubes manufactured using ShAPE \cite{shaped1}. During ShAPE, a rotating die impinges on a stationary billet placed within an extrusion container equipped with a coaxial mandrel. At the interface between the billet and the die, the billet is heated and plasticized by both the shear forces applied and by the resulting frictional heat. As the die moves into the plasticized billet material, the material emerges from a cavity in the die to form a tube extrudate. ShAPE process parameters are comprised of data streams such as tool rotation rate and tool traverse rate (feed rate), which result in specific extrusion temperatures, forces, torques, and power.

The data used in this study was developed by \cite{WHALEN2021699}, which resulted in manufactured AA7075 tubes. The authors then obtained tube coupons at different locations and subjected them to T5 and T6 heat treatment before finally testing them to determine their ultimate tensile strength (UTS), yield strength (YS) and elongation. The coupons were also imaged using a scanning electron microscope to obtain the fore-scatter and back-scatter images of the microstructure of the samples. Of the several process-microstructure-property data streams available from the original study, in this work we narrowed our scope to consider a single ShAPE processing parameter, the feed rate, a single resulting material property, the UTS, and the back-scatter modality of SEM images for analysis in this project. We also account for the post-ShAPE heat treatment (T5, T6 tempers) as it can strongly influence material properties for ShAPE materials and those from other processes more broadly. We illustrate the interplay of process parameters, material properties, and SEM microstructure in Figure \ref{fig:triangle} by generating conditioned SEM images corresponding to either specific feed rates and temper at which the samples were manufactured or UTS.

\subsection{Generative Adversarial Networks}
\begin{figure}[t]
    \centering
    \includegraphics[width=0.9\columnwidth]{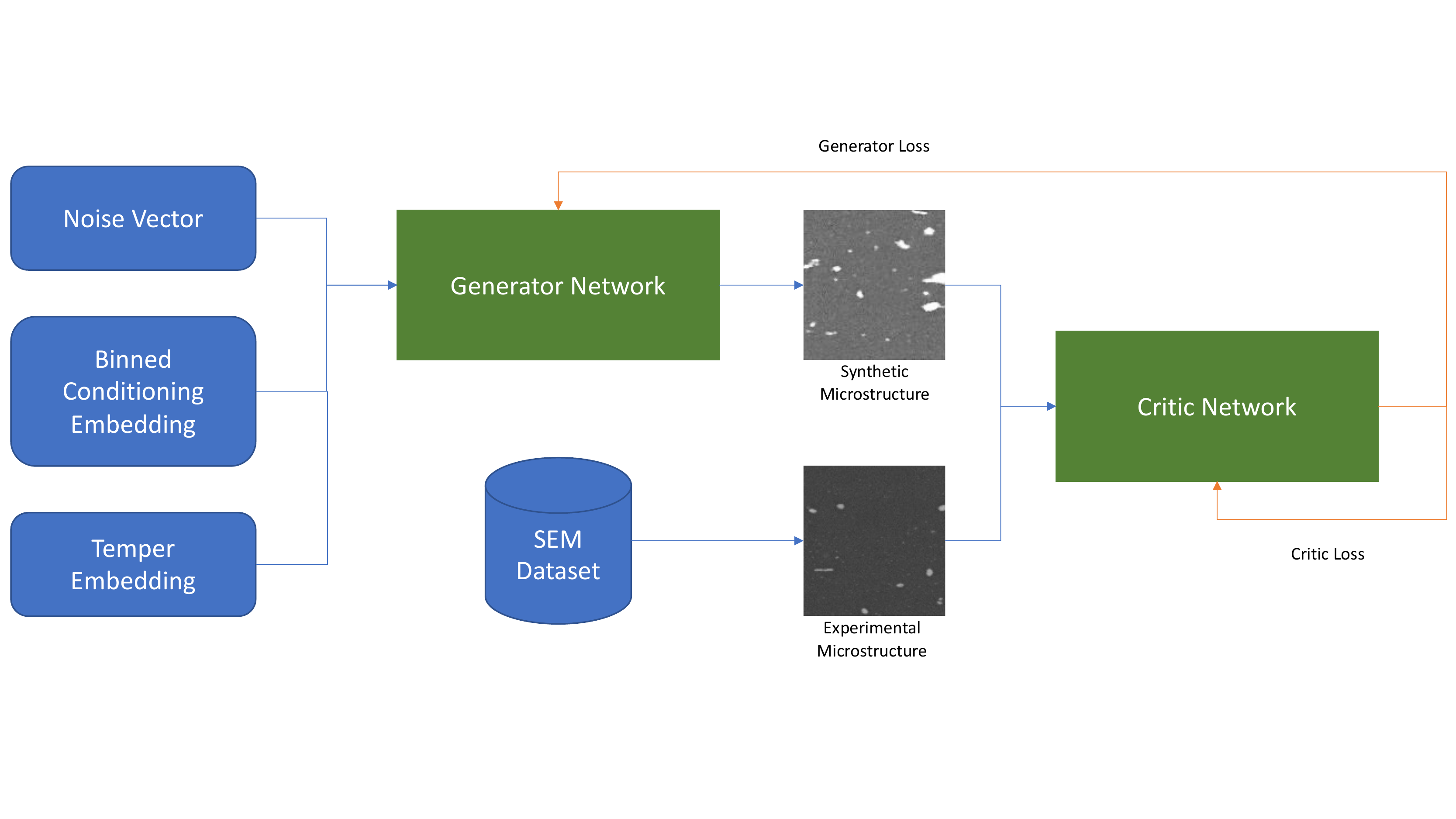}
    \caption{A high-level overview of our GAN model's component and data flow. Blue elements are associated with input data while green elements are trainable neural network components. Blue arrows constitute the model's forward pass whereas orange arrows represent backpropagation updates to the model's parameters.}
    \label{figure:gan}
\end{figure}

Generative Adversarial Networks (GANs)  \cite{goodfellow2014generative} are designed to model a training data distribution by way of an adversarial game played by two neural networks: a generator and a discriminator network. The generator takes as input a noise vector, typically sampled from a standard normal distribution, and uses this entropy source to generate a unique data sample - for our purposes, an image. The discriminator takes as input a data sample and predicts whether it is a real sample from the training set, or whether it is a fake sample produced by the generator. The networks are adversarial in that they are optimized to fool one another, with the goal of producing a generator which can sample data indistinguishable from the original training distribution. 

Since their introduction, GANs have undergone multiple extensions and improvements: some aimed to stabilize their training dynamics, others to improve their generation quality, and others still to allow GANs to incorporate new sources of domain information. Conditional GANs (CGANs) \cite{mirza2014conditional} provide side channel information, typically class labels, to both the generator and discriminator networks. This facilitates sampling from sub-distributions of the original training data. This approach was improved upon by Auxiliary Classifier GANs (AC-GANs) \cite{odena2017acgans}, which only provide side channel information to the generator and task the discriminator with learning to determine both authenticity \textit{and} the side channel information of a given sample.

Another avenue for GAN improvement arrived with the introduction of the Wasserstein GAN \cite{arjovsky2017wgan}, which introduced a new loss formulation based on the Wasserstein distance \cite{kantorovich1939mathematical} to improve training stability. This formulation changed the discriminator to produce an authenticity \textit{score} rather than a simple real-fake classification, lending these particular discriminators the title of ``critics". Later work in this vein introduced a gradient penalty to the WGAN's loss function (WGAN-GP) \cite{gulrajani2017improved-wgan}, improving training stability yet further.

\subsection{Microstructure Topological Feature Analysis}

\emph{Persistent homology} is a popular tool from topological data analysis used to study the shape of data \cite{edelsbrunner2000topological}.
In particular, \emph{sublevel set persistent homology} is a technique frequently applied to grayscale image data to study the variation in intensity of patches (neighboring pixels) in an image\footnote{In our analysis, we use one-dimensional homology $H_1$.}. 
In this context, a $m \times n$ persistence image (PI) is often used to represent the persistence of topological features 
of the data across scaling \cite{adams2017persistence}.
It summarizes the creation time (or \emph{birth}) on the horizontal axis and persistence time on the vertical axis as topological features.

Recent work \cite{emerson2022toptemp} has demonstrated that persistence images can act as powerful and robust feature descriptors for microstructures like those collected from ShAPE AA7075 tubes - specifically microstructures well characterized by precipitate intensity/contrast and precipitate density/distribution.  Given the interpretability, generalizability, and noise robustness provided by these methods, we leverage them in evaluating the fidelity between experimental and synthetic SEM microstructures. By comparing PIs from experimental and synthetic SEM images, we are able to visualize and discuss similarities and differences between the two image types (real and generated) in a feature space that preserves domain knowledge~\cite{emerson2022toptemp}. The relationships between PIs are visualized by using Principal Component Analysis (PCA) to perform dimensionality reduction so that both groups can be visualized together in 3-dimensions.

\section{Experimental Methods}
\subsection{SEM Image Dataset}
We took as our GANs' input data a collection of SEM images of 32 AA7075 ShAPE tubes manufactured using 32 different process conditions followed by T5 and T6 heat treatments. The SEM microstructure images are large (2560$\times$1920 pixels), grayscale, and were taken at 500$\times$ magnification. The most important feature of interest in the AA7075 ShAPE tube SEM images are the intermetallic precipitates which are seen as lighter particles with varying morphology and topology. However, 32 samples is far too little to train a deep learning model, but due to the relatively small precipitate size in these images and the lack of large salient precipitates we cropped each large SEM image into smaller chips of size 128$\times$128. The chips were partially overlapped and yielded a training set of 437,000 images which we used in our experiments.

Our data preprocessing choices were enabled by the precipitate form and structure itself. In the ShAPE AA7075 tube microstructures, the critical microstructures can be observed over a fairly limited, local spatial extent. We arrived at our crop size by identifying the smallest chip that a human expert would be capable to effectively evaluating. This choice further depends on the magnification level. If, for another manufacturing process, the entire SEM image is needed for microstructure evaluation, it may be more practical to learn to generate descriptions or \emph{features} of the SEM images rather than the entire image when only a small number of training samples are available.  Future efforts will explore ways to numerically determine these parameter choices.

Even with a larger number of images to work with, we still have a very small number of overall experiments, which are the source of the experimental parameter and material property values we use for conditioning information - namely ShAPE feed rate and UTS. Initial experiments where we conditioned our models on normalized scalar values did not perform well. We suspect this was because the conditioning information was simply too sparse, and that the models overfit and were unable to extrapolate beyond this small experiment set.

As has been done elsewhere in the literature \cite{ahmad2018rvc,ammar2019geometric, truong2021differential, workman2016horizon} we relaxed the desired regression problem into a classification problem by discretizing the feasible range of values into categorical variables. In doing so we transformed the conditioning information (either process parameters values or property values) into three categories: ``low'', ``mid'', and ``hi''. These bins divided the scalars observed in our experiments into lower, middle, and upper thirds. This is a coarse binning, but one that could still significantly accelerate manufacturing research if predicted accurately by providing guidance about possible SEM microstructures that can be obtained when a samples is manufactured in a specific regime that demonstrates UTS above or below a certain user-specified number. 

Additionally, nine of the ShAPE experiments produced extrudates which did not undergo further T5 or T6 tempering. Since UTS was not available for the ShAPE tubes that were not heat treated further, we only include these ``as extruded'' experimental microstructures when learning the effects of ShAPE feed rate on resulting microstructures. This temper imbalance, combined with imbalances between our ``low'', ``mid'', and ``hi'' conditioning labels, gave us the per-setting experiment counts depicted in Table \ref{tab:exp-bin-dists}.

\begin{table}[t]
\centering
\begin{tabular}{|l|l|l|}
\hline
\textbf{Temper} & \textbf{\begin{tabular}[c]{@{}l@{}}Feed Rate\\ Label Counts\end{tabular}} & \textbf{\begin{tabular}[c]{@{}l@{}}Ultimate Tensile Strength\\ Label Counts\end{tabular}} \\ \hline
T5              & 9 / 8 / 15                                                                   & 11 / 7 / 3                                                                                   \\ \hline
T6              & 9 / 8 / 15                                                                   & 12 / 5 / 4                                                                                   \\ \hline
\end{tabular}

\caption{The number of experiments within our ``low'', ``mid'', and ``hi'' label binnings for feed rate and ultimate tensile strength (UTS) conditions for both T5 and T6 temper treatments. As a process parameter selected before an experiment begins, feed rate is independent of tempering. Bin values correspond, respectively, to lower, middle, and upper thirds of observed values.}
\label{tab:exp-bin-dists}
\end{table}

\subsection{GAN Architecture and Training}
For our generative models we used Wasserstein GANs with gradient penalty \cite{gulrajani2017improved-wgan} and an auxiliary classifier, abbreviated ACWGAN-GP. Its generator module takes as input the concatenation of three inputs: a 100-dimensional noise vector sampled from a standard normal distribution, as well as learned, 20-dimensional dense embedding vectors for two tempering conditions (corresponding to T5 and T6 treatments) and similarly learned 20-dimensional ``low'', ``mid'', and ``hi' label embeddings for either a manufacturing experimental parameter or a material property. From these inputs, which we concatenate along the feature dimension, the generator produces a grayscale image. The GAN's critic module takes as input a grayscale image and produces separate scores for: the WGAN critic score (scalar), temper classification (two logits), and the correct binning label for the experimental parameter or material property (three logits). We trained two GANs in our experiments: one conditioned on feed rate and one conditioned on desired UTS. 

We trained all GANs for 400 epochs over our training set of SEM crops, with a batch size of 64 and the AdamW optimizer \cite{loshchilov2017decoupled}. During training we upweighed each model's gradient penalty by a factor of 10 and the critic's classification losses by a factor of 5. We found these values gave an effective training speed as well as diversity and quality of synthetic images.

Our approach is similar to that of \cite{iyer2019conditional}, with differences emerging from our data and our particular framing of the conditional generation problem. The ShAPE data afforded us a significantly greater number of training samples (437,000 vs approx. 7,000), each of which has a ``simpler'' visual structure (ex. Figures \ref{fig:uts-samples} and \ref{fig:feedrate-samples}) than samples from the Ultra High Carbon Steel DataBase (UHCSDB) \cite{decost2017uhcsdb}. Our work examined two tempering conditions (T5 and T6) instead of five, though we explored joint conditioning over both temper and parameter or material property conditions. That our work explores conditional generation over process parameters and material properties is novel in itself.

\section{Results and Analysis}
\subsection{Synthetic Image Quality}
\begin{figure}[t]
    \centering
    \includegraphics[width=0.9\columnwidth]{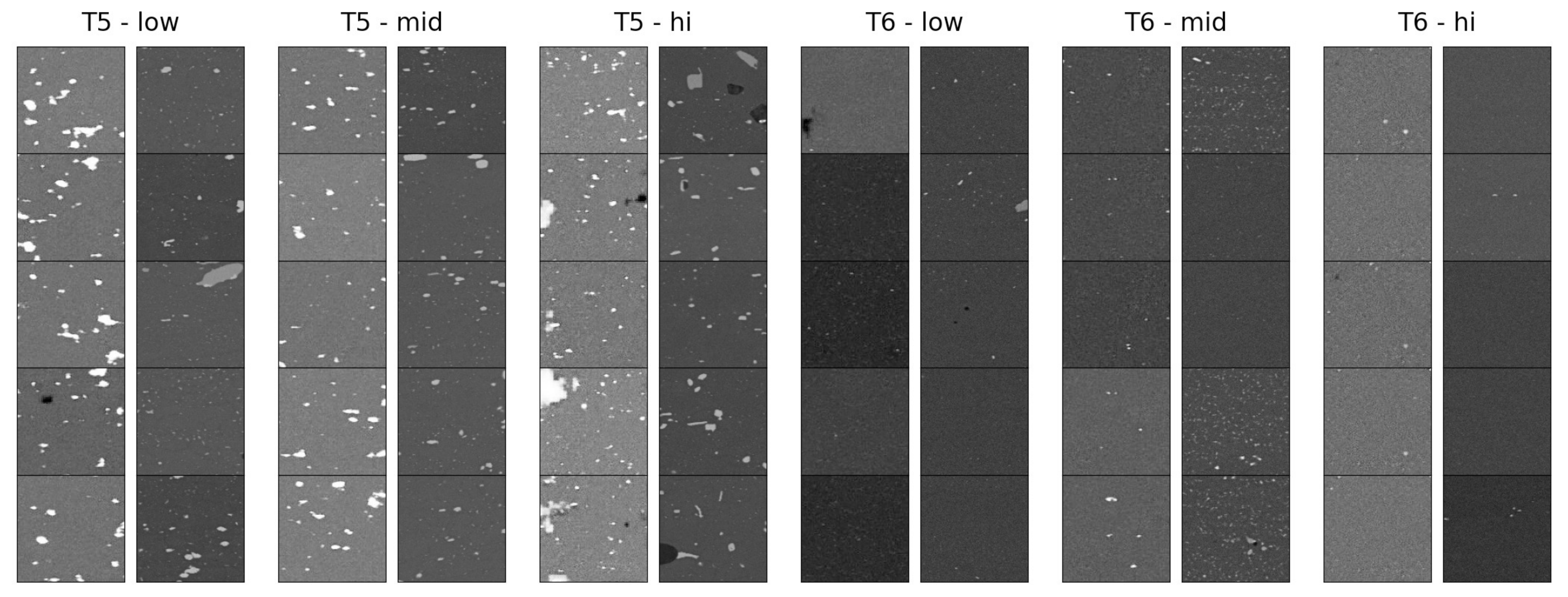}

    \caption{Paired columns of synthetic (left) and experimental (right) ultimate tensile strength-conditioned SEM images under T5 and T6 temper settings.}
    \label{fig:uts-samples}
\end{figure}

\begin{figure}[t]
    \centering
    \includegraphics[width=0.9\columnwidth]{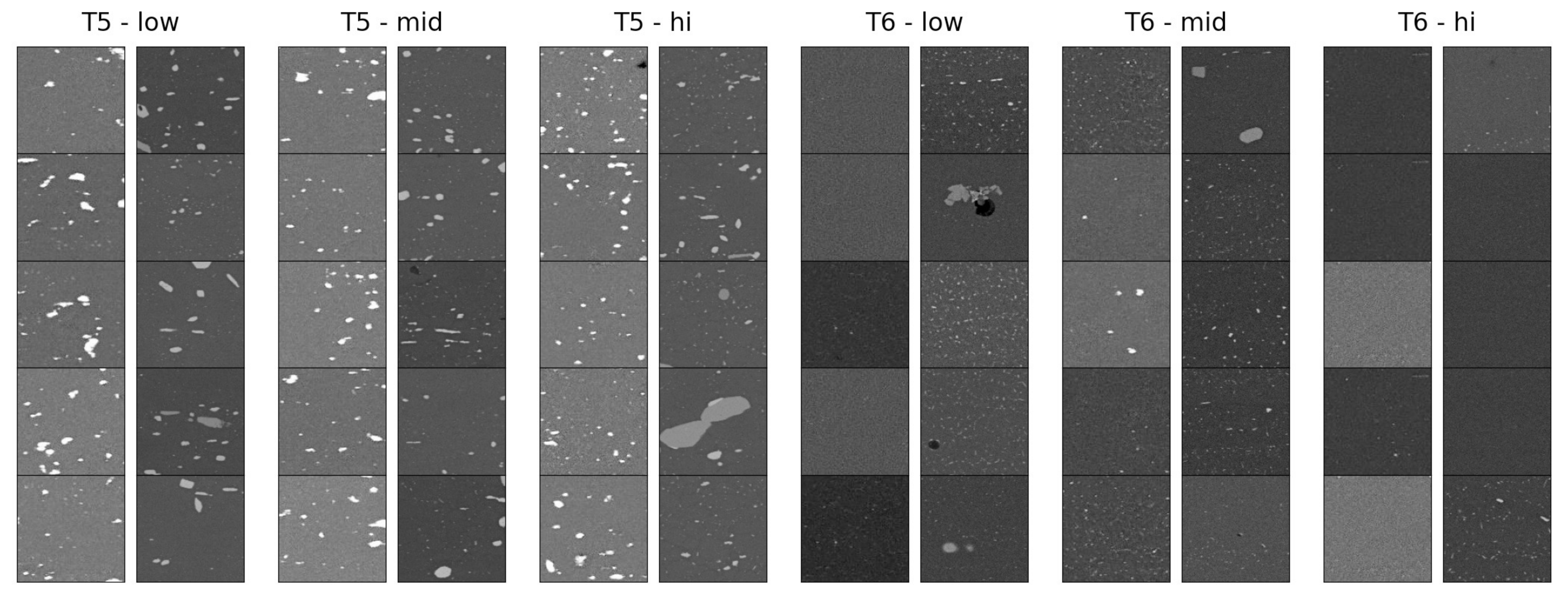}

    \caption{Paired columns of synthetic (left) and experimental (right) feed rate-conditioned SEM images under T5 and T6 temper settings.}
    \label{fig:feedrate-samples}
\end{figure}

Figures \ref{fig:uts-samples} and \ref{fig:feedrate-samples} show synthetic images conditioned on T5 and T6 temper conditions jointly with ``low'', ``mid'', and ``hi'' ultimate tensile strength (UTS) or feed rate, respectively. We pair the synthetic images with experimental ones for comparison. Among some notable differences between the experimental and synthetic images, we see that exceptionally large precipitates appear more artifacted and less frequently in synthetic images than in experimental ones, especially under the T6 condition.

Despite the heavy imbalance of binned labels across experiments shown in Table \ref{tab:exp-bin-dists}, we saw synthetic T5 and T6 samples consistently display high and low degrees of sample diversity, respectively, regardless of the number of experiments associated with a given label bin. We suspected this was because T6-tempered microstructures tend to have far fewer salient precipitates against a background of tiny, visually near-random precipitates. This produces less variation throughout an uncropped SEM image, and by extension less variation among our training crops. This hypothesis is also consistent with the lack of mode collapse (a common issue encountered when training GANs wherein the model learns to produce a single plausible output at the expense of output diversity) in the T5 setting, where the presence of many larger, irregular precipitates means crops from within a single SEM image will be highly diverse. Additionally, the artifacting of large precipitates in the T5 condition could benefit from differentiable data augmentation \cite{zhao2020differentiableaugmentation} as a way to upsample rare phenomenon in our experimental data. Incorporating recent GAN techniques targeting efficient use of unbinned, scalar conditioning labels \cite{zheng2021continuous} could further improve performance by removing the need for a discretely-conditioned, categorical latent space in favor of more continuous ones, able to better leverage what is presently unwieldy conditioning information. Recent GAN regularization techniques based on consistency \cite{zhao2021ganconsistency} and separating the discrimination and classification tasks \cite{li2021triplegan} could supplement these approaches more generally.

While visual quality is an important indicator of our GANs' performances, it doesn't tell the full story. In order to be useful for experimental design and property analysis, our models must produce images which are physically meaningful, not just visually plausible. Our remaining evaluations are focused on understanding the degree to which our GANs' image distributions align with our experimental ShAPE data.

\subsection{Topological Fidelity Experiments}
\begin{figure}[t]
    \centering
     \includegraphics[width=0.6\columnwidth]{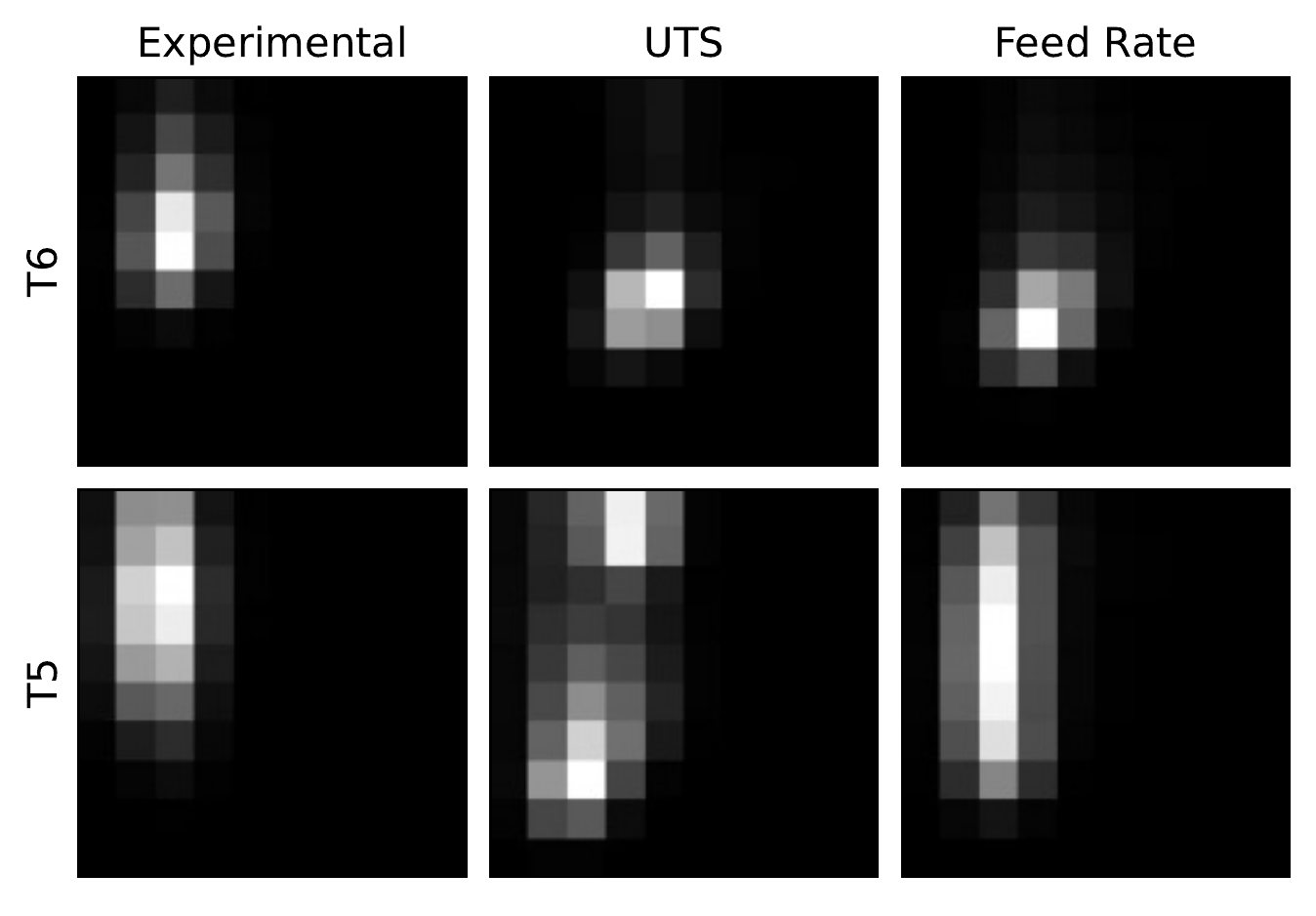}

    \caption{Average persistence images (PIs) over experimental (top) and synthetic (bottom) ShAPE SEM chips. The PIs we use detect the number of ``holes'' in an image, a statistic that has been shown to aligned with scientifically salient features. PIs that are similar indicate that the number and scale of ``holes'' between two images is similar. We compare across synthetic imagery conditioned on either feed rate or ultimate tensile strength (UTS) and across T5 and T6 temper conditions.}
    \label{fig:persistence-images}
\end{figure}

We average the $10\times 10$ PIs of experimental and synthetic images: Figure \ref{fig:persistence-images} shows average PIs derived from experimental samples as well as all of our GANs. Briefly, each PI is constructed by gradually increasing a threshold which is used to capture spatially contiguous pixels with intensity up to the threshold. As the threshold increases, more pixel clusters (precipitates in our case) will fall below the threshold and be ``born''. This is captured by the horizontal axis: pixels further to the right capture pixel clusters that are "born" under a later threshold - in our case, it captures brighter precipitates. The PIs also capture how long a given cluster ``lives'' before eventually merging with other clusters as the threshold continues to increase. This merging is based on the proximity and relative intensity of these neighboring clusters and so captures their spatial distribution. Pixels higher along the vertical axis represent precipitates that have longer ``persistence''. For example, we would expect a microstructure with a blend of low and high-brightness precipitates to have a PI with clusters of pixels on both the left and right; we would also expect the PI of a microstructure with a few large, distant, bright precipitates to have more pixels concentrated towards the top of the image than a microstructure with many small precipitates evenly sprinkled throughout, even if the brightness of these precipitates were the same in both cases.

At a high level, we can see that in our synthetic and experimental images, the general shape of the persistence pattern is preserved; however, the synthetic images do not have precise agreement with the experimental image distributions. Interestingly, we see that our synthetic images produce an overly concentrated persistence pattern in the T5 case, but an overly diffuse persistence pattern in the T6 case. Since these are \textit{average} persistence images, we conclude that our distribution of synthetic T5 images exhibit insufficient levels of structural variation (mild mode collapse), whereas the T6 distribution exhibits unrealistically \textit{high} levels of variation. The overly ``tall'' T6 persistence patterns are consistent with our observation that our GANs are able to rely too heavily on producing a blanket of tiny, noisy precipitates compared to experimental references. Partial mode collapse is also visible in Figures \ref{fig:uts-samples} and \ref{fig:feedrate-samples} in the form of synthetic samples of a consistent, high average brightness. Some experimental samples have a comparable degree of brightness but the synthetic image distribution fails to capture the experimental diversity in both precipitate and background brightness. This problem might be alleviated by denoising the backgrounds of experimental images which would free the generator and discriminator from needing to model blankets of noise, especially in the T6 setting. Such denoising would not introduce additional training signal, but reducing the degree of confounding or task-irrelevant visual information could aid the GAN optimization process.

At the same time, the synthetic persistence patterns are still close to their experimental counterparts. The exception would be the PI for the synthetic images corresponding to T6 condition conditioned on UTS, which exhibits an oddly bimodal persistence pattern indicative of partial mode collapse. These results are consistent with GANs which are trained to mimic the visual distribution without access to topological or physical regularization, which could better align them with experimental SEM data.

\begin{figure}[t]
    \centering
    \includegraphics[width=0.7\columnwidth]{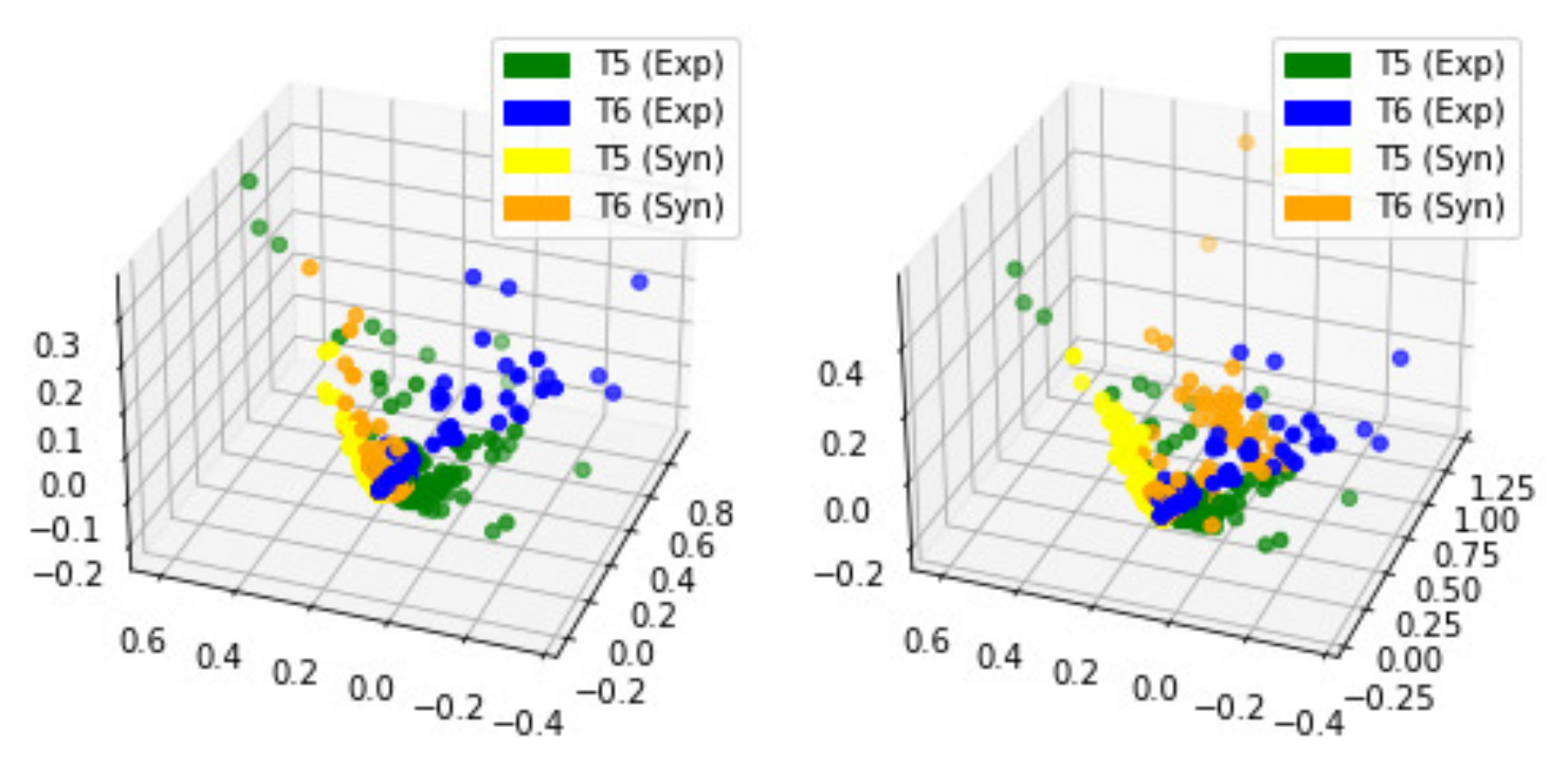}
    
    \caption{We project persistence images into a 3D space using principal component analysis to to reduce the dimenions of our experimental data, with synthetic images sampled from our models conditioned on either ultimate tensile strength (left) or feed rate (right).}
    \label{fig:pi-pca}
\end{figure}

Equipped with PIs as features, we perform principal component analysis (PCA) \cite{pearson1901liii} to reduce the dimensions of experimental imagery so that it can be visualized. We can then inspect the alignment between our experimental and synthetic data. Figure \ref{fig:pi-pca} visualizes the projection of our synthetic and experimental PI images onto the experimentally-fit components for either feed rate or ultimate tensile strength (UTS)-based conditioning. We can see in both cases that our synthetic PIs do not overlap well with the experimental PIs for both T5 and T6 temper conditions; indeed, they appear nearly orthogonal in component space. We also note that there is exceptionally less agreement between T5 and T6 projections for the synthetic, feed rate-conditioning case than all other settings. This is consistent with the average PIs in Figure \ref{fig:persistence-images} and with the visual samples in Figures \ref{fig:feedrate-samples} and \ref{fig:uts-samples}. The average PIs reflect a similar dynamic due to the feed rate-conditioned T6 PIs having having an unrealistically high spread of intensities compared to all other settings, whereas the image samples show that the feed rate-conditioned T6 model is poor at capturing the larger precipitates in the experimental data - even compared to our UTS-conditioned model.

These PCA results provide evidence that our model generators learn a manifold which is \textit{visually} similar to the ground truth data distribution, at least in the T5 setting, but which is \textit{topologically} quite distinct. This is counter to intuition around GANs where in theory the discriminator will pressure the generator into matching the true data distribution, in all is aspects, over time. However, in general the history of GAN research shows the myriad ways in which this process can fail to live up to its theoretical potential: unstable optimization dynamics, imbalanced generator-discriminator training, and other difficulties can produce unsuccessful or only partial alignment between the generator and ground truth data distributions. While issues of imbalanced generator-critic steps are largely accounted for by our Wasserstein GAN architecture, we experimented with more powerful critics to verify that this issue was not caused by underparameterization. Specifically, we replace our several-layer convolutional network with a ResNet-18 \cite{he2016deep} either randomly initialized or pretrained on ImageNet \cite{russakovsky2015imagenet} and see \textit{degraded} performance compared to our initial, smaller architecture. We hypothesize that this discrepancy is not due to lack of model capacity but rather due to challenges in the optimization process: the generator can only be pressured into adopting topologically realistic data if the critic itself can distinguish disparate topological features. Incorporate explicit topological regularization, or a second small critic network providing feedback on derived persistence images, would be a useful avenue for future work to enforce a more physically plausible synthetic image distribution.

\begin{figure}[t]
    \centering
    \includegraphics[width=0.9\columnwidth]{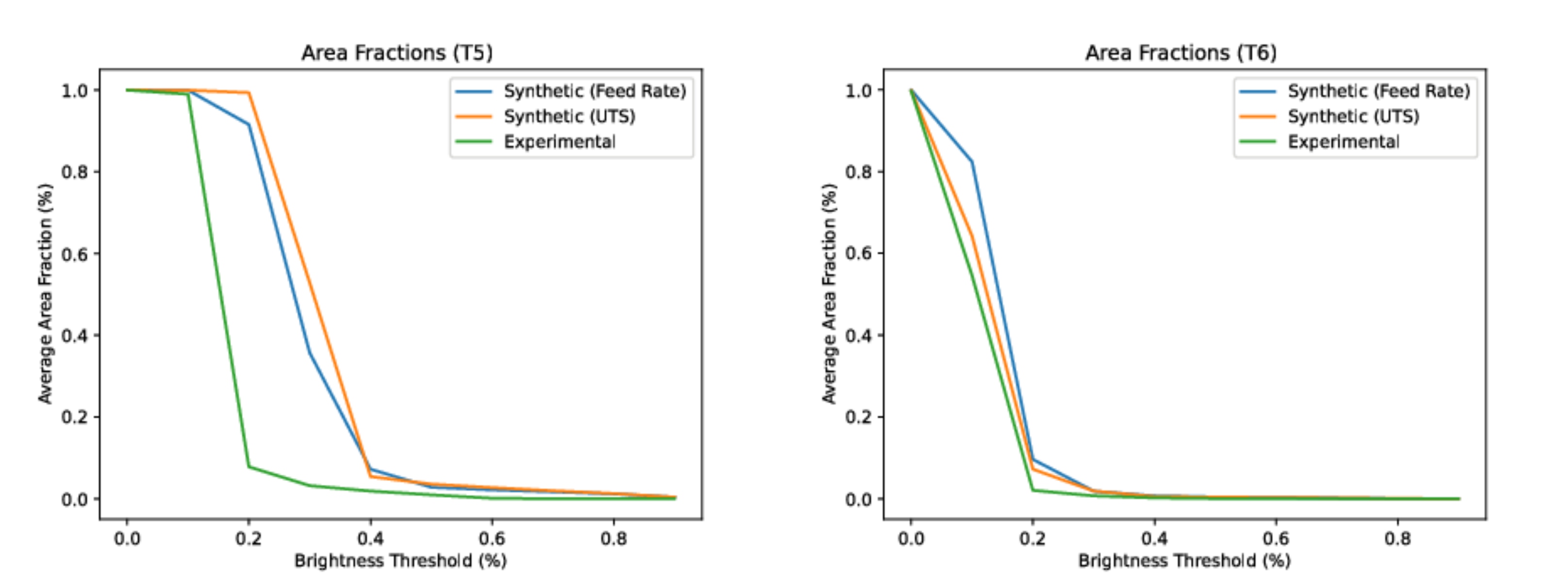}
    
    \caption{Area fraction statistics for our synthetic and experimental data across T5 and T6 temper conditions. Fractions are calculated using 1,000 images randomly selected from each condition. We performed mean centering and scaling to $[0,1]$ within each condition for the sake of comparison.}
    \label{fig:area_fractions}
\end{figure}

Finally, we examined the area fractions of bright pixels across synthetic and experimental microstructures; this is done because such area fractions are a common heuristic used in material science applications to evaluate precipitate density in microstructures. Because we were interested in comparing area fractions across conditions in order to ascertain the prevalence of bright precipitates against the microstructures' darker backgrounds, we performed feature scaling within each condition such that images share a center mean and pixel intensities lie in the range $[0,1]$. In order to measure the area fraction of a given image, we then set a threshold $t \in [0,1]$ and calculated the percentange of pixels with intensity $i \geq t$. We used ten evenly spaced thresholds in $[0,1]$, with the results captured by Figure \ref{fig:area_fractions}.

The results show that the variation in mean area fraction between synthetic and experimental T5 microstructures is less than 10 percent. This implies that in the generated SEM images, the precipitates occupy an area fraction that roughly coincides with that of the experimental images. This is not the case for T6 microstructures, where we observe that experimental datasets have consistently higher area coverage of darker to mid-tone precipitates. This is consistent with both our visual observations in Figures \ref{fig:uts-samples} and \ref{fig:feedrate-samples} and in the persistence images in Figure \ref{fig:persistence-images}. We suspect this phenomenon is caused by the lower precipitate density in our experimental T6 data: when each image has low precipitate density, our models are able to minimize their loss functions by producing disproportionately large amounts of background noise in all cases. Regularizing our models by including loss terms for persistance image patterns or for area fraction scores could alleviate this problem and would be productive avenues for future work.

These results indicate that, despite an intuitive visual plausibility, albeit one that doesn't fully capture the natural variation of experimental data, our models are learning a distribution over SEM images that is \textit{topologically} distinct from experimental data. The generated images show precipitate morphology and distribution which result in similar area fractions as those observed in T5 experimental microstructures, though less so for T6 microstructures. Given the utility of topological features for characterizing SEM imagery \cite{emerson2022toptemp}, we pose improving this alignment as a useful avenue for future work. However, these limitations are unsurprising for a generative model trained on largely visual stimuli alongside a simple conditional embedding; the visual quality of our synthetic data indicates that it is possible to train generative deep learning models even on the relatively small data scale afforded by the advanced manufacturing domain, and our ability to pinpoint deficiencies in terms of latent space or topological feature alignment allows us to see actionable steps towards more physically realistic, deployment-ready development of this approach.

\section{Conclusion}
This work takes a critical first step towards a functional machine learning-accelerated advanced manufacturing experimental pipeline. We trained multiple conditional Wasserstein GANs (ACWGAN-GPs) on SEM microstructure image crops derived from AA7075 manufactured using the advanced ShAPE process. This is an advance over prior work which focuses on \textit{unconditional} SEM generation for steels, marking a step towards a generative system that scientists can query to predict how either process parameters or properties impact a material microstructure. We observe that our synthetic images are visually plausible, though with some visual artifacting of rare precipitate phenomena. Additionally, we observe through topological methods of inquiry that our synthetic image distributions do not uniformly align with experimental SEM images. Specifically, we see dissimilarity between experimental and synthetic images in ways consistent with the small number of unique experiments present in most advanced manufacturing datasets. In future work, we propose exploring two avenues to address these limitations: differentiable data augmentation and recent developments in GAN regularization as a way to better leverage limited advanced manufacturing data and to increase model sample efficiency, as well as topological and physical regularization to encourage the GANs to produce synthetic data which expresses even higher fidelity to experimental data distributions.

\section*{Acknowledgements}

This work was performed using the resources available at the Pacific Northwest National Laboratory (PNNL) and funded by the Mathematics for Artificial Reasoning in Science (MARS) Initiative as a Laboratory Directed Research and Development Project. KSK thanks Scott Whalen, Md. Reza-E-Rabby, Tianhao Wang, Scott Taysom and Timothy Roosendaal for the ShAPE AA7075 tubes synthesis and property data. KSK also appreciates Woongjo Choi for their contributions to data arrangement; Tianhao Wang, Xiaolong Ma, and Alan Schemer-Kohrn for developing the SEM images of the ShAPE AA7075 tubes; and Luke Gosink, Elizabeth Jurrus, and Sam Chatterjee for their support and advice on this project. PNNL is a multi-program national laboratory operated by Battelle Memorial Institute for the U.S. Department of Energy under contract DE-AC05-76RL01830.

\section*{Conflict of Interest Statement}
On behalf of all authors, the corresponding author states that there is no conflict of interest.

\bibliography{references.bib}




\end{document}